\documentclass[letterpaper, 10 pt, conference]{ieeeconf}  

\IEEEoverridecommandlockouts                              

\usepackage{graphics} 
\usepackage{epsfig} 
\usepackage{mathptmx} 
\usepackage{times} 
\usepackage{amsmath} 
\usepackage{amssymb}  
\usepackage{todonotes}
\usepackage{cancel}
\usepackage{caption}
\usepackage{booktabs}

\definecolor{MyDarkBlue}{rgb}{0,0.08,1}
\definecolor{MyDarkGreen}{rgb}{0.02,0.6,0.02}

\usepackage{hyperref}

\usepackage{color}
\newcommand{\revise}[1]{\textcolor{black}{#1}} 

\begin{document}
\title{\LARGE \bf Terrain-Aware Quadrupedal Locomotion via Reinforcement Learning}

\author{Haojie Shi$^{1,2*}$, Qingxu Zhu$^{1*}$, Lei Han$^{1}$, Wanchao Chi$^{1}$, Tingguang Li$^{1\dag}$ and Max Q.-H. Meng$^{3\dag}$, \textit{Fellow, IEEE} 
\thanks{ This work was partially supported by National Key R\&D program of China
        with Grant No. 2019YFB1312400, Hong Kong RGC CRF grant C4063-18G, Hong Kong RGC GRF grant \#14211420 and Hong Kong RGC TRS grant T42-409/18-R awarded to Max Q.-H. Meng.  \textit{(Corresponding authors: Tingguang Li,Max Q.-H. Meng)}}
\thanks{$*$ Equal Contribution.}%
\thanks{$^{1}$ Haojie Shi, Qingxu Zhu, Lei Han, Wanchao Chi and Tingguang Li are affiliated with Tencent Robotics X, China, (email: {\tt\footnotesize{haojieshi, qingxuzhu,lxhan,wanchaochi,teaganli}@tencent.com})}%
\thanks{$^{2}$ Haojie Shi is from the Chinese University of Hong Kong, and this work was done during internship at Tencent Robotics X Lab.}%
\thanks{$^3$ Max Q.-H. Meng is with Shenzhen Key Laboratory of Robotics Perception and Intelligence and the Department of Electronic and Electrical Engineering at Southern University of Science and Technology in Shenzhen, China. He is a Professor Emeritus in the Department of Electronic Engineering at The Chinese University of Hong Kong in Hong Kong and was a Professor in the Department of Electrical and Computer Engineering at the University of Alberta in Canada. (email: {\tt\footnotesize max.meng@ieee.org})}%
}

\maketitle
\begin{abstract}
In nature, legged animals have developed the ability to adapt to challenging terrains through perception, allowing them to plan safe body and foot trajectories in advance, which leads to safe and energy-efficient locomotion. Inspired by this observation, we present a novel approach to train a Deep Neural Network (DNN) policy that integrates proprioceptive and exteroceptive states with a parameterized trajectory generator for quadruped robots to traverse rough terrains. Our key idea is to use a DNN policy that can modify the parameters of the trajectory generator, such as foot height and frequency, to adapt to different terrains. To encourage the robot to step on safe regions and save energy consumption, we propose foot terrain reward and lifting foot height reward, respectively. By incorporating these rewards, our method can learn a safer and more efficient terrain-aware locomotion policy that can move a quadruped robot flexibly in any direction. To evaluate the effectiveness of our approach, we conduct simulation experiments on challenging terrains, including stairs, stepping stones, and poles. The simulation results demonstrate that our approach can successfully direct the robot to traverse such tough terrains in any direction. Furthermore, we validate our method on a real legged robot, which learns to traverse stepping stones with gaps over $25.5$cm.

\end{abstract}

\section{INTRODUCTION}
	
In recent years, there has been significant progress in the field of quadrupedal locomotion. Both model-based control methods \cite{bellicoso2017dynamic,bledt2018cheetah,carius2018trajectory,carius2019trajectory,di2018dynamic,fankhauser2018robust,carpentier2018multicontact,aceituno2017simultaneous,winkler2018gait,farshidian2017efficient} and learning-based methods \cite{da2020learning,lee2020learning,peng2018deepmimic,rudin2022learning,yang2020multi,yang2022fast,li2023learning,han2023lifelike} have demonstrated advantages in robust locomotion control over various terrains. However, most of these methods have primarily focused on proprioceptive information and neglected the importance of exteroceptive information.

In fact, the ability to perceive the environment is crucial for legged animals to navigate safely and efficiently through different environments. For example, legged animals can jump over large gaps with awareness of the gap locations to avoid falling. Furthermore, legged animals can adjust their foot height based on the terrain. For instance, they lift their feet higher on stair-like terrains than on flat surfaces, thus conserving energy. Inspired by this, researchers have attempted to incorporate exteroceptive information, mainly visual information, to design safer locomotion controllers for challenging terrains.

\footnotetext[1]{We provide a video to show the results in  \href{https://youtu.be/mjIxrDC_QtQ}{$youtu.be/mjIxrDC_QtQ$}.}

\begin{figure}[t]
	\vspace{-.3cm}
	\centering
	\includegraphics[width=0.48\textwidth]{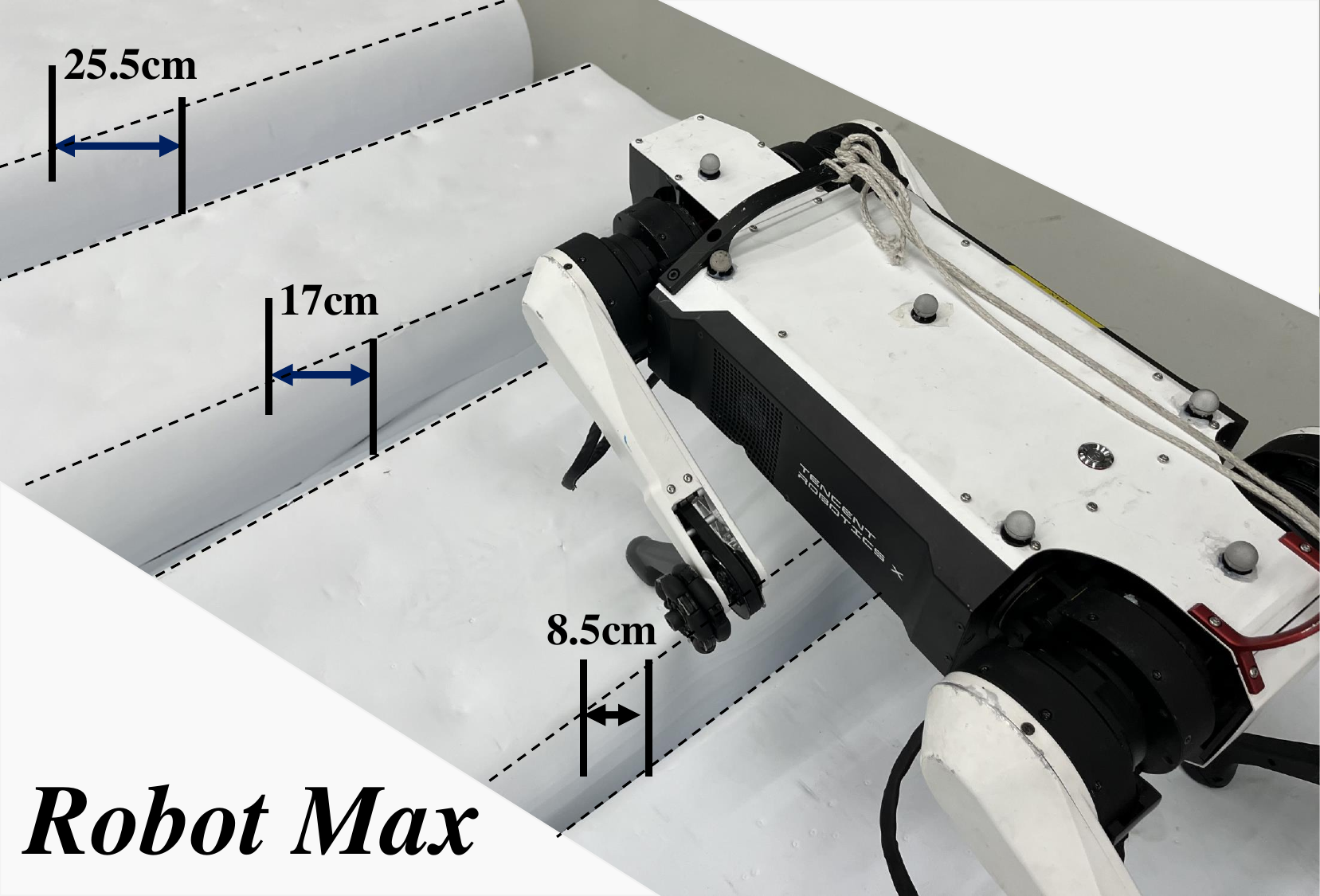}
	\caption{The quadrupedal robot \emph{Max} traversing the stepping stones.}
	\label{fig:setting}
	\vspace{-.6cm}
\end{figure}

Several recent works \cite{margolis2021learning,miki2022learning,fu2022coupling,yang2021learning,yu2021visual} have taken advantage of deep neural networks to process visual information for robotic locomotion. Typically, these works adopt a hierarchical approach, whereby a vision-based deep neural network policy is used to generate a reference trajectory, which is then tracked by a model-based controller \cite{yu2021visual,margolis2021learning}. This method decouples trajectory planning and control, facilitating the training of safe locomotion policies. However, the hierarchical structure requires significant prior knowledge of the robot's dynamics model and manual effort to design the low-level controller, either with model predictive control \cite{camacho2013model} or whole-body control \cite{bellicoso2017dynamic}. In contrast, our work trains a DNN policy with a parameterized trajectory generator, without the need for any model-based controller. Notably, few purely learning-based approaches have been successful in traversing challenging discrete terrains, such as stepping stones and poles, without a model-based controller.

Furthermore, prior works have paid little attention to how exteroceptive information can be used to generate more efficient trajectory plans that minimize energy consumption. In contrast, our work incorporates an exteroceptive reward signal that encourages the robot to lower its lifting foot height in different terrains, thereby reducing energy consumption.

The goal of our work is to develop a framework that integrates both proprioceptive and exteroceptive information to generate a safer and more energy-efficient trajectory plan directly via deep reinforcement learning (DRL). Following the approach in \cite{iscen2018policies}, we design a parameterized trajectory generator to produce a nominal trajectory for a quadrupedal robot. By incorporating perceived terrain information, the DRL policy is trained to adjust the parameters of the trajectory generator, such as foot height and frequency, to adapt to different terrains.
In contrast to prior works \cite{lee2020learning,miki2022learning}, our approach utilizes a more flexible trajectory generator that can adjust foot height to save energy consumption based on visual information. Furthermore, we propose a new reward term, the foot terrain reward, which encourages the robot to step on safe regions based on visual input. By training our policy in a DRL setting, we can achieve a safer and more energy-efficient locomotion strategy without requiring a model-based control policy.

We evaluate the effectiveness of our approach through a series of experiments conducted in both simulation and on a real quadruped robot. In simulation, our robot safely traverses block terrains, stairs with heights up to 15cm, stepping stones with gaps up to 25cm, and poles with diameters ranging from 10cm to 15cm. The simulation results demonstrate that our approach can generate safer and more efficient trajectory plans for the quadruped robot to travel in any direction. In addition, physical experiments with our real robot confirm the effectiveness of our approach, as the robot successfully traverses stepping stones with large gaps up to 25cm. These results demonstrate the potential of our approach to enable safe and efficient locomotion in real-world scenarios.

In conclusion, the main contributions of this paper are:
\begin{enumerate}
	\item We propose a deep reinforcement learning framework with a parameterized trajectory generator that integrates both the proprioceptive and exteroceptive information. \revise{Our DNN policy can produce residual joint angles and adjust the trajectory generator's parameters from terrain information.}
	\item \revise{We propose foot terrain reward to encourage the robot to step on safe regions, and lifting foot height reward to lower the robot's lifting foot height to save energy consumption.} It can make a safer and more efficient trajectory plan for the quadrupedal robot to traverse various difficult terrains in any direction.
	\item We evaluate our approach both in simulation and the real world across several tough terrains such as stairs, stepping stones, and poles, and the experiment results show the effectiveness of our method.
\end{enumerate}
\section{RELATED WORK}

\subsection{Model-based control}

Model-based control methods \cite{bellicoso2017dynamic,bledt2018cheetah,carius2018trajectory,carius2019trajectory,di2018dynamic,fankhauser2018robust} are widely used for quadrupedal locomotion due to their ability to derive optimal control forces based on the dynamic model of the robot. These methods use optimal control theory such as quadratic programming and model predictive control \cite{camacho2013model} to compute the optimal force inputs. For example, \cite{di2018dynamic} uses Model Predictive Control(MPC) to compute force inputs that track the reference trajectory derived from a simplified dynamics model of the quadrupedal robot. \cite{bellicoso2017dynamic} proposes a whole-body controller that optimizes the whole-body motion and contact forces to track the reference trajectory. Although these methods demonstrate robust control performance, they heavily rely on the accuracy of the dynamics model and require sophisticated trajectory planners to produce agile locomotion skills.

\subsection{Learning-based control}

Benefiting from the development of deep reinforcement learning algorithms \cite{schulman2017proximal,schulman2015trust}, learning-based control methods have made tremendous breakthroughs in quadrupedal locomotion \cite{da2020learning,lee2020learning,peng2018deepmimic,rudin2022learning,yang2020multi,yang2022fast,shi2022reinforcement}. These methods leverage the nonlinearity of deep neural networks to learn the control policy directly by maximizing the accumulated rewards when interacting with the simulation environment. This framework can learn the control policy automatically without manual effort to design the reference trajectory. Furthermore, through imitation learning, they can learn more agile locomotion skills such as those seen in nature \cite{peng2018deepmimic}. Multi-expert adaptive skills, such as trotting, steering, and fall recovery, can also be learned using the gating neural network \cite{yang2020multi}. However, there remains a significant gap between the simulation environment and the real world, making it difficult to deploy the motion skills learned in simulation to the real robot. Although researchers have proposed methods such as dynamics randomization to narrow this gap \cite{peng2018sim}, transferring the policy learned in simulation to the real world remains an open problem for learning-based methods.

\subsection{Visual Locomotion}
Previous methods for locomotion have primarily relied on proprioceptive information, which is sufficient for safe movement on continuous terrains but may fail on discrete or tough terrains, such as stepping stones. With the key role of perception information in allowing robots to perceive their environment and avoid dangerous terrains, researchers are now paying greater attention to visual locomotion. For example, \cite{miki2022learning} has developed a robust visual locomotion policy that allows robots to traverse terrains in the wild by creating an attention-based recurrent encoder to encode visual information. To tackle discrete terrains such as stepping stones, \cite{yu2021visual,margolis2021learning} have used a hierarchical strategy that trains deep neural network policies to produce reference trajectories and model-based control policies to track them. While these methods have been successful in developing safe visual locomotion policies for tough terrains, the use of a hierarchical strategy that combines model-based control increases the manual effort required compared to purely learning-based methods. Instead, our work proposes a purely learning-based framework that can directly learn safe and energy-efficient locomotion policies for tough discrete terrains.

\begin{figure*}[ht]
	\vspace{-.3cm}
	\centering
	\includegraphics[width=\textwidth]{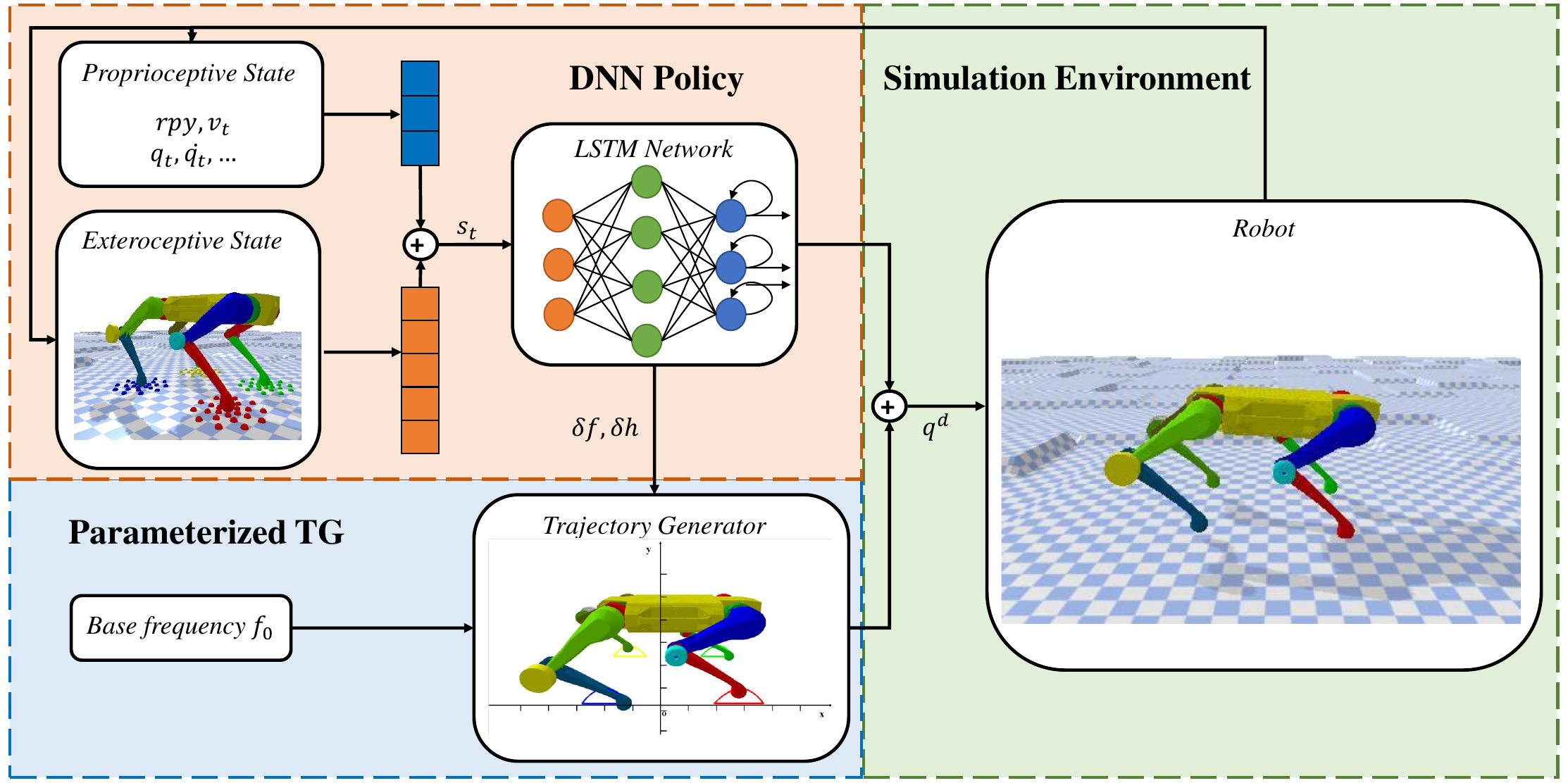}
	\caption{Our proposed framework contains the DNN policy module and the parameterized trajectory generator module to control the robot. Our DNN policy integrates both the proprioceptive and terrain information, and deploys the LSTM \cite{hochreiter1997long} network to output the residual of desired joint position and also adjust the parameters of the trajectory generator. }
	\label{fig:framework}
	\vspace{-.4cm}
\end{figure*}

\section{METHODOLOGY}
    \subsection{Framework}
    Our framework contains a deep neural network policy and a Trajectory Generator (TG) which is parameterized by the DNN. 
     The TG is a pre-defined open-loop controller that can produce nominal reference control signals and generate cyclic motions. In the field of learning-based approaches for locomotion, such trajectory generators can speed up policy training since they can generate reference trajectories \cite{lee2020learning}. However, its parameters, i.e., the amplitude, and the lifting foot height, are fixed, which limits its capability for complex situations.
    To make the TG adapt to the terrain and robust to external perturbations, we exploit the DNN policy to parameterize the TG. The DNN takes both proprioceptive and exteroceptive information as input, and outputs the residual joint angles and adjusts the TG's frequency and lifting foot height parameters. The adjusted TG can generate a nominal reference trajectory, and the residual joint angles can further adjust the desired joint angles to adapt to different terrains. Moreover, the DNN policy can lower the robot's lifting foot height in different terrains to save energy consumption by adjusting the TG's lifting foot height parameter. Figure~\ref{fig:framework} illustrates our proposed framework.
    
    \subsection{Trajectory Generator}
    The reference foot trajectory is generated by the parameterized trajectory generator, which provides prior knowledge for the quadrupedal robot controller and accelerates the training of reinforcement learning \cite{iscen2018policies}. In our proposed framework, we use the trotting gait, which is a cyclic motion of the four legs, and we represent the timing of the motion cycle by a phase $p\in [0, 2\pi]$. Given the phase of the left front leg $p_{LF}$, we can compute the phase of the other legs as $p_{RR}=p_{LF}$ and $p_{RF}=p_{LR}=p_{LF}+\pi$. Therefore, we only need to consider the phase of the left front leg in the following discussion.
    The current phase $p_t$ is computed by accumulating the current phase frequency $f_t$, that is:
    \begin{equation}
        p_t = p_{t-1}+f_t*T
    \end{equation}
    Where $T$ is the time step. Given the constant base frequency $f_{base}$ and the residual frequency $\delta f$ from the DNN policy, the current phase frequency is computed as:
    \begin{equation}
        f_{t}=f_{base}+\delta f
    \end{equation}
     Similar to \cite{lee2020learning}, after deriving the current phase of each leg $p_i$, the desired residual position of each leg is computed as:
    \begin{gather}
        q = |p_i/\pi|\\
        d_l = -(2q^3-3q^2+1)+0.5\\
        dx = c_x*d_l \label{eq:dx} \\
        dy = c_y*d_l \label{eq:dy} \\
        t=2*p_i/\pi \\
         dz=\left\{
        \begin{aligned}
         & 0  , & p_i \leq 0  \\
        & (-2t^3+3t^2)*(H+\delta h)  ,& 0< p_i \leq \pi/2  \\
        & (2t^3-3t^2+1)*(H+\delta h)  ,& p_i > \pi/2  \\
        \end{aligned}
        \right.
    \end{gather}
    Where $H$ is the predefined foot height, $\delta h$ is adjusted by the DNN policy, \revise{ $d_l\in[-0.5,0.5] $ is the normalized phase term, and $c_x, c_y$ are the displacement of one footstep in XY plane. Given the pre-defined footstep length $L$ and the command direction $\theta$, we can derive $c_x, c_y$ as:
    \begin{gather}
        c_x=Lcos\theta\\ c_y=Lsin\theta
    \end{gather}
     }
    After deriving the desired residual position of each foot, the actual desired position can be computed by:
    \begin{equation}
        pos^d = [dx+X,dy+Y,dz+Z]
    \end{equation}
    Where [X, Y, Z] is our robot's constant base foot position set as [0, 0, -0.28]. Moreover, the desired position is represented in the horizontal frames of each leg~\cite{barasuol2013reactive}. Then we can further compute the joint reference angles of TG by Inverse Kinematics (IK). The trajectory generator module in Fig.~\ref{fig:framework} sketches how our TG functions.
    
    Compared with the TG in \cite{lee2020learning}, our parameterized TG is capable of adjusting the foot height by adding $\delta h$ from the DNN policy, and it also provides the movement in XY plane $dx,dy$, where the TG in \cite{lee2020learning} only considers the movement in the Z-axis.

    \subsection{Reinforcement Learning}
    We train our DNN policy via a deep reinforcement learning algorithm, which learns the optimal policy to maximize the cumulative rewards when interacting with the environment. In this work, we employ the PPO algorithm~\cite{schulman2017proximal} to train our policy, which is one of the state-of-the-art online learning algorithms.
    \subsubsection{Observation}
    Our observation $s_t$ contains both proprioceptive and exteroceptive state. The proprioceptive state consists of moving direction command $c_t = [cos(\theta),sin(\theta),0]$ with $\theta$ from 0 to $2\pi$, body rotation and angular velocity, joint angles and velocity, the history of action output, frequency factors of the trajectory generator, and the history of desired foot position.
    
    Concerning the exteroceptive state, we simplify the terrain information by a height map around each foot of the robot, which is similar to \cite{miki2022learning}. For each foot $f_i$, we sample $n$ points $p_{ij}$, $j = \{1,2,...,n\}$ along circles of different radius centered at $f_i$, and derive the height of the terrains at such points $h_{ij}$, $j=\{1,2,...,n\}$. We finally compute our exteroceptive state information by regulating the height with the foot height, that is, $z_{ij}=f_{ij}-h_{ij}$. The sampled points can be seen in the exteroceptive state module of Fig.~\ref{fig:framework}.
    \subsubsection{Policy and Action}
    Our policy first encoded both the proprioceptive and exteroceptive state into a latent space and used LSTM network \cite{hochreiter1997long} to leverage temporal information from history. Given the DNN policy $\pi_\theta$ and observation $s_t$, the action $a_t$ can be computed by $a_t=\pi_\theta(s_t)$. As for action space, our action has 14 dimensions. The first two dimensions $a_{t,1-2}$ adjust the frequency and foot height of the trajectory generator $tg$, and the following 12 dimensions $a_{t,3-14}$ are the residual desired joint angles. Finally the desired joint angles are computed via $q^d = a_{t,3-14}+tg(a_{t,1-2})$. With the desired joint angles, a PID controller computes each joint's torque output $\tau_t$. 
    \subsubsection{Reward}
    Our proposed reward function includes six terms: the velocity within command reward ($r_v$), the velocity out of command reward ($r_{vo}$), the energy reward ($r_{tau}$), the foot terrain reward ($r_{terrain}$), the lifting foot height reward ($r_{height}$), and the smoothness reward. The first two rewards encourage the agent to move along the desired direction. The energy reward is designed to minimize energy consumption. The foot terrain reward encourages the robot to avoid areas that pose a risk of falling, such as big gaps. The lifting foot height reward encourages the robot to lower the height at which it lifts its feet in different terrains to save energy. The smoothness reward is designed to promote a natural and smooth gait. These rewards are computed based on various inputs, including the body velocity ($v_t\in \mathbb{R}^3$), the desired velocity range ($[v_{l},v_{h}]$), the torques of each joint ($\tau_t\in\mathbb{R}^{12}$), the joint angles ($q_t\in\mathbb{R}^{12}$) and velocities ($q_{v,t}\in\mathbb{R}^{12}$), the terrain height around each foot ($z_{i,j}$), the height threshold ($H_{thre}$) for detecting discrete and unsafe areas, the height difference in the neighboring environment ($\Delta H$), and the height threshold ($F_{thre}$) for lifting the feet over the terrain. The rewards are computed as below:
    \begin{equation}
        r_v=\left\{
        \begin{aligned}
        & 1.0  , & v_l\leq v_t^Tc_t\leq v_h, \\
        & e^{-2{(v_t^Tc_t-v_h)}^2}  , &  v_t > v_h,\\
        & e^{-2{(v_t^Tc_t-v_l)}^2}  , &  v_t < v_l,
        \end{aligned}
        \right.
    \end{equation}
    \begin{equation}
        r_{vo}=e^{-1.5{(v_t^Tv_t-v_t^Tc_t)}^2}
    \end{equation}
    \begin{equation}
        r_{tau}=-\tau_t^Tq_{v,t}
    \end{equation}
    \begin{equation}
        r_{terrain,i}=\left\{
        \begin{aligned}
         0 & , \text{$f_i$ is swing or $max(z_{i,j})-min(z_i,j)>H_{thre}$ } & \\
         -1 & ,  elsewise &\\
        \end{aligned}
        \right.\label{eq:terrain}
    \end{equation}
    \begin{equation}
        r_{height,i} = -max(H+\delta h-\Delta H-F_{thre},0)\label{eq:height}
    \end{equation}
    \begin{equation}
        r_{smooth} = e^{-0.5{(q_t-q_{t-1})}^T(q_t-q_{t-1})}
    \end{equation}
    
    \revise{From Eq.\ref{eq:terrain}, we can see that the foot terrain reward punishes the agent when it steps into the gap where its terrain height is much lower than safe ground regions so that it can encourage the policy to generate more safe plans. And Eq.\ref{eq:terrain} shows that the lifting foot height reward will be negative only when the robot lifts its foot much higher than the terrain. In this way, the policy will efficiently learn to adapt the robot's lifting foot height in different terrains to save energy consumption.}
    \begin{figure}[ht]
	\vspace{-.3cm}
	\centering
	\includegraphics[width=0.48\textwidth]{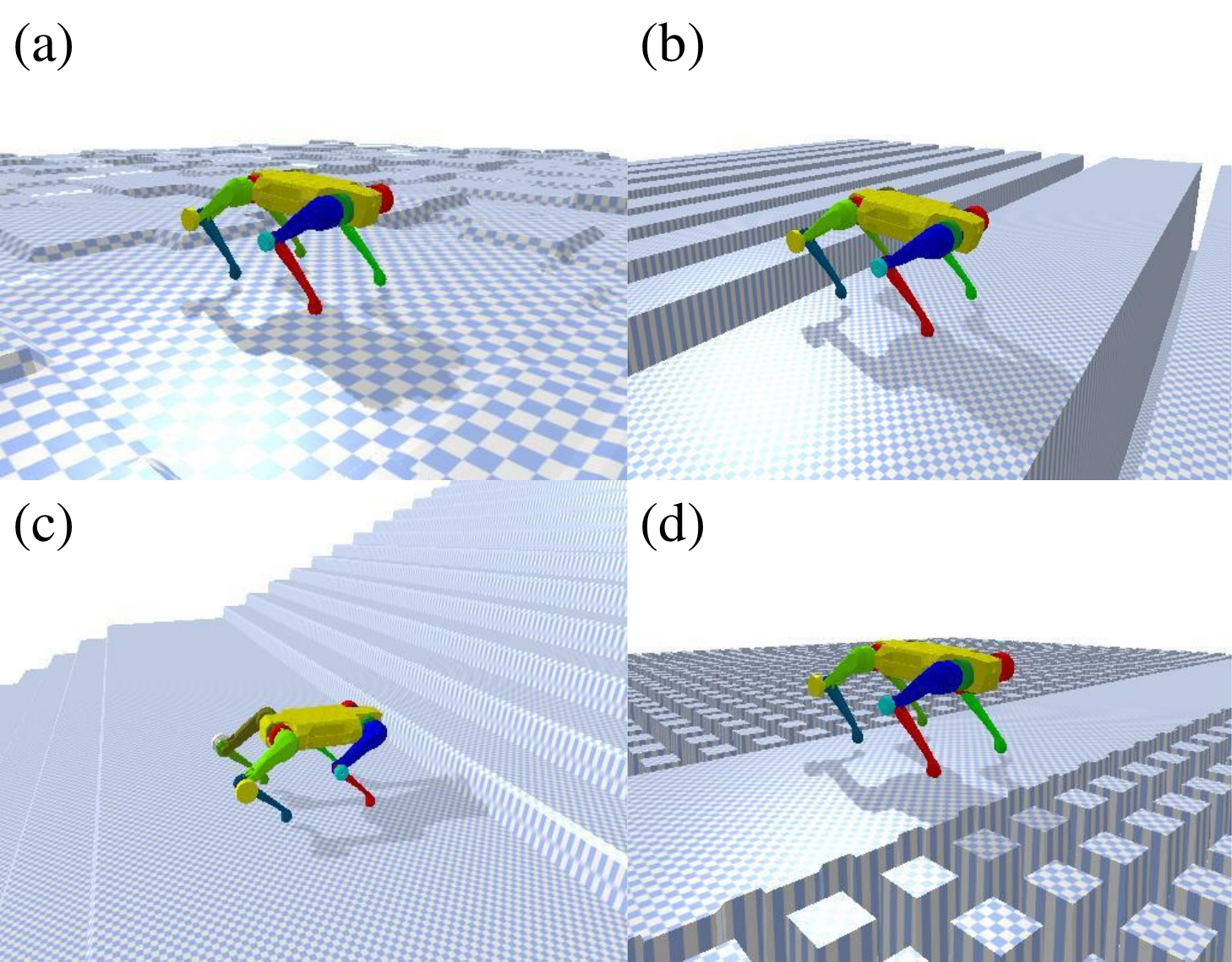}
	\caption{Simulation environments. (a) Block terrains. (b) Stepping Stones. (c) Stairs. (d) Poles.}
	\label{fig:env}
    	\vspace{-.4cm}
    \end{figure}
    
\section{EXPERIMENTS}
    To evaluate the effectiveness of our methods, we conducted experiments in both the PyBullet simulation environment \cite{coumans2016pybullet} and on a real quadrupedal robot. Our experiments aim to answer the following key questions:
    \begin{enumerate}
	\item Can our approach generate safer plans for the robot when navigating challenging and discrete terrains?
	\item Is our approach able to generate plans that lower the robot's lifting foot height to save energy consumption?
	\item Is our approach robust when deployed to the real robot in difficult terrains like traversing the stepping stone?
	\end{enumerate}
\subsection{Simulation Experiments}
    As shown in Fig.~\ref{fig:env}, our simulation environments contain four types of terrains: blocks, stepping stones, stairs, and poles. For perception, we sample $[1, 6, 8, 10]$ points from the circles with the radius of $[0cm, 3cm, 6cm, 10cm]$, respectively, centered at each foot as shown in Fig.~\ref{fig:framework}.
    
    To validate our methods, we demonstrate the success of our trained policy in directing the robot to safely traverse terrains in any direction. Additionally, we compare our approach with other learning-based methods, including an approach that uses only proprioceptive information for end-to-end learning, and another approach that uses a fixed-height trajectory generator from \cite{lee2020learning} that cannot adjust foot height or produce movement in the XY plane.
    \revise{
    \begin{table}[]
\centering
\caption{The mean and standard deviation of maximum travel distance of all the policies in different terrains. It is tested in six different directions and run three times in each direction. The maximum travel distance is limited by five meters due to the size of the map. }
\label{tab:compare}
\resizebox{0.5\textwidth}{!}{%
\begin{tabular}{@{}cccc@{}}
\toprule
             &Ours           & Fixed-height TG    & w/o perception                \\ \midrule
Block  & \textbf{4.983\textpm0.038}    & 4.911\textpm0.140    & 4.948\textpm0.104  \\
Stair  & 4.951\textpm0.114    & 4.859\textpm0.383      & \textbf{4.960\textpm0.092}  \\
Stepping Stone  & \textbf{4.954\textpm0.070}   & 1.782\textpm1.054    & 1.539\textpm0.856  \\
Poles  & \textbf{4.930\textpm0.099}    & 1.924\textpm1.132    & 1.906\textpm0.925  \\
 \bottomrule
\end{tabular}%
}
\vspace{-4mm}
\end{table}
}
    \begin{figure*}[ht]
    	\vspace{-.3cm}
    	\centering
    	\includegraphics[width=\textwidth]{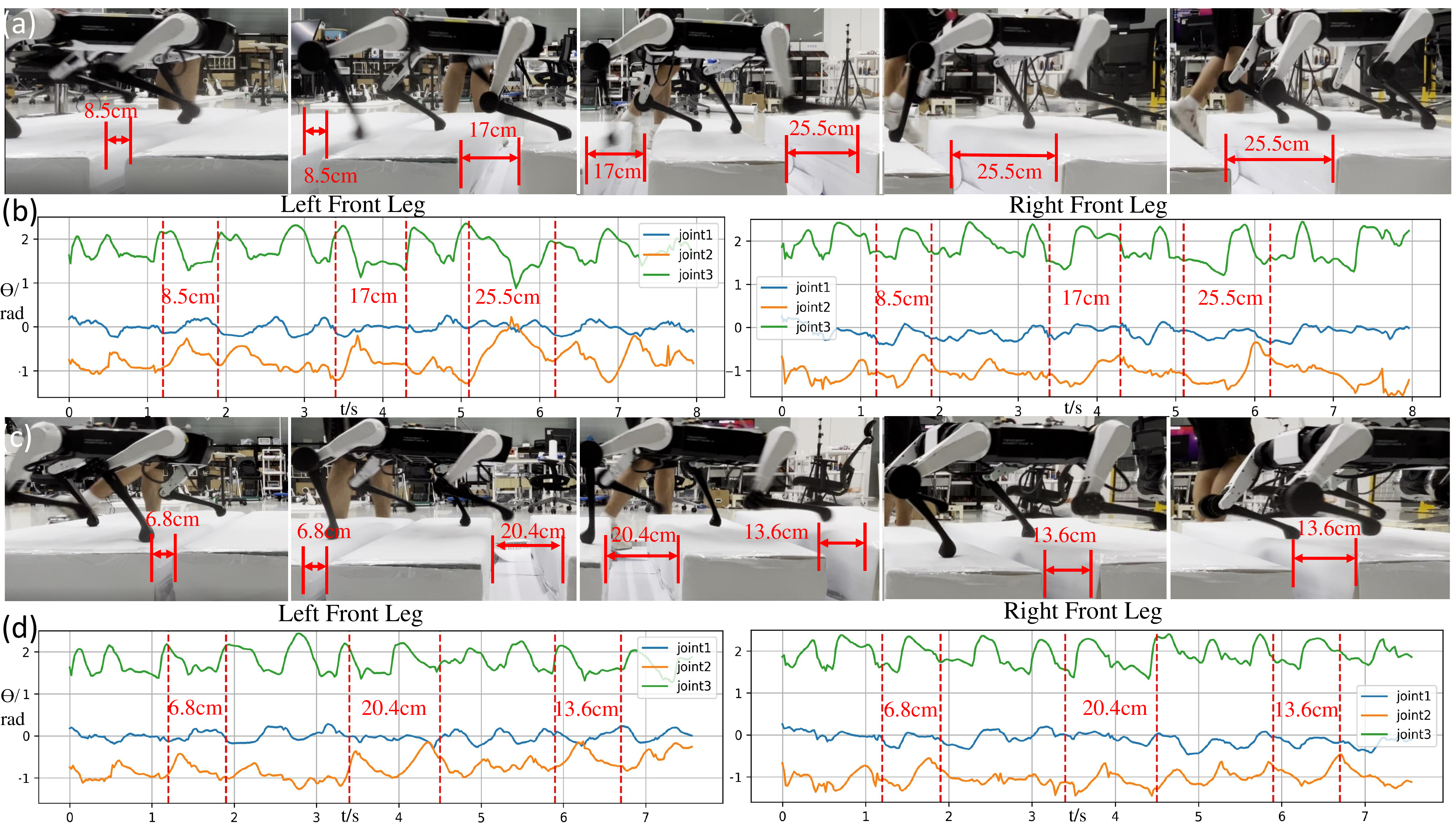}
    	\caption{Our robot Max traverses gaps of various widths. (a) Traversing gaps with widths of [8.5cm,17cm,25.5cm]. (b) The desired joint angles produced by our policy for the left front leg and the right front leg in the first experiment. The red lines indicate when the robot steps over the corresponding gaps. (c) Traversing gaps with widths of [6.8cm, 20.4cm, 13.6cm]. (d) The same as (b). For (b), (d), the amplitude of the second joint adjusts depending on the gap width, with a significant increase when crossing gaps with widths of 25.5 cm and a decrease when crossing gaps  with widths of 8.5 cm.}
    	\label{fig:exp}
    	\vspace{-.4cm}
    \end{figure*}
            \begin{figure}[ht]
    	\vspace{-.3cm}
    	\centering
    	\includegraphics[width=0.5\textwidth]{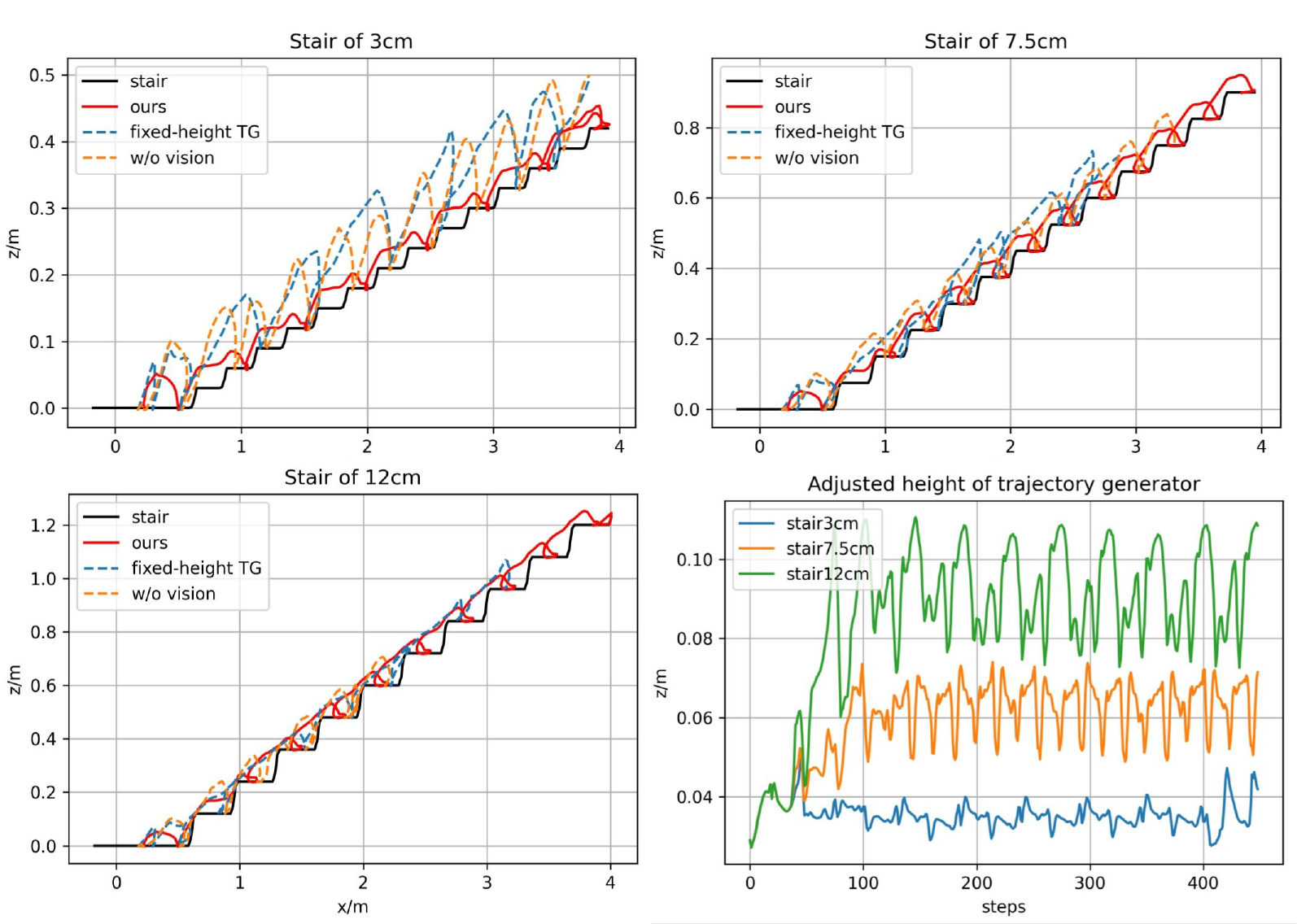}
    	\caption{The foot trajectory is produced by all the methods when the robot climbs the stairs of various heights. (a) Stair with a height of 3cm. (b) Stair with a height of 7.5cm. (c) Stair with a height of 12cm. (d) The adjusted foot height $H+\delta h$ of the TG in stairs of different heights.}
    	\label{fig:stair}
    	\vspace{-.4cm}
    \end{figure}
    
   \revise{To answer the first question, we conduct experiments to compare the maximum travel distance of the above methods in all the terrains so as to evaluate the safety metric.} We do not limit the time steps so that we can compare the safety of the policy without considering the difference in the walking speed of different policies. To justify if the policy can direct the robot in any direction in each environment, we sample 6 different direction commands from $[0,\pi/4,-\pi/4,-\pi,-5\pi/4,-3\pi/4]$, where 0 direction means walking forward, and $-\pi$ means walking backward. We conduct three random experiments for each direction sample and compute its \revise{ mean and standard deviation of maximum travel distance}. The final results are shown in Table \ref{tab:compare}. From the table, we can see that for the block and stair terrains, it is easy to train a safe policy even without perception since these terrains are continuous. However, for discrete terrains like stepping stones and poles, it is observed that the policy without perception fails to travel through them since the perception information is the key information for such problems. Meanwhile, the policy with the fixed-height TG in \cite{lee2020learning} also fails. The reason may be that the trajectory generator in \cite{lee2020learning} does not contain the displacement in the XY plane \revise{, which may encourage the robot to move forward the desired direction.} Moreover, our experiments find that in such a setting, it is easy to learn a policy that can only walk forward in the continuous terrains like blocks while stopping when facing discrete terrains like stepping stones. On the other hand, our proposed TG can produce movement signals in the XY plane by computing $dx,dy$ in Eq. \ref{eq:dx}, \ref{eq:dy}, and it will lead the policy to try to move forward when facing tough terrains. Hence, our approach can produce safety trajectory plans in continuous and discrete terrains.
    
    Regarding the second question, we aim to determine whether our proposed approach can generate energy-efficient plans by adjusting the lifting foot height of the robot when facing stairs of varying heights. We hypothesize that a good policy should learn to raise the robot's feet only as high as necessary to overcome stairs of different heights, rather than as high as possible for all stairs. To evaluate this, we compare the foot trajectory produced by our method with those produced by the other methods when the robot climbs stairs of different heights ranging from 3cm to 12cm.  From Fig.~\ref{fig:stair}, we can see that our policy can adjust the lifting foot height based on different stairs while others raise the robot's feet much higher when facing lower stairs. Furthermore, we plot the actual height parameter $H+\delta h$ of our TG that is adjusted by DNN policy in different stairs in Fig.~\ref{fig:stair} (d). From the plot, we can see that our DNN policy learns to adapt the foot height of TG on various stairs. \revise{ In particular, for stairs with a height of 3cm, our policy generates foot trajectory plans that allow the robot to lower its lifting foot height to around 3cm, while other methods force the robot to lift its foot over 10cm. These results demonstrate the superior energy efficiency of our proposed approach.}
    
\subsection{Physical Experiments}
    For the physical experiments, we implemented our algorithm on our real robot Max. Focusing on evaluating our proposed framework, we simplified the physical experiment's setting by using motion capture. However, future work could be done to derive terrain information using Lidar sensors. We used motion capture to obtain the relative coordinates between the robot's position and the terrain map, and then used the offline map to derive the height map around the robot's feet. It should be noted that motion capture was only used to obtain perception information, and not to estimate any proprioceptive state. To bridge the sim2real gap, we employ dynamics randomization to introduce variations in the environment, such as changes in ground friction and perception noise.
    To mitigate the noisy velocity estimation in the real robot, we exclude body velocity from the proprioceptive state during policy training

    The results can be seen in the supplemented video. In Fig.~\ref{fig:exp}, we plot two experiments where our robot traverses stepping stones of different gaps. The widths of gaps in the first experiment are [8.5cm ,17.0cm ,25.5cm], and in the second experiment are [6.8cm, 20.4cm, 13.6cm]. From the results, we can see that our robot can succeed in traversing gaps of various widths safely. Fig.~\ref{fig:exp} also plots the desired joint angles produced by our DNN policy for the left front leg and right front leg. The plots show that the angle of joint two will be adapted according to different gaps. \revise{ In particular, Fig.~\ref{fig:exp} (b) and (d) demonstrates that the amplitude of the second joint adjusts depending on the gap width, with a significant increase in amplitude when crossing larger gaps.}

\section{LIMITATIONS AND CONCLUSION}
    Although our approach is capable of generating safe and efficient trajectory plans by adapting to the environment, it has a shortcoming in proving safety guarantees when traversing tough discrete terrains. The first challenge is to formulate safety constraints in tough discrete terrains to prevent falling. Then, exploring how to train the deep reinforcement learning policy under the safety constraints is necessary. Additionally, to implement our algorithm in the wild as in \cite{miki2022learning}, generating the height map by the lidar scan is also needed. In conclusion, our proposed end-to-end learning algorithm with a parameterized trajectory generator integrates proprioceptive and exteroceptive states to produce safe and efficient trajectory plans in tough terrains. The deep neural network policy learns to adjust the trajectory generator's parameters to generate the plans. The experimental results prove the effectiveness of our algorithm in generating safe and efficient plans in terrains such as blocks, stairs, stepping stones, and poles.

\newpage
{\small
	\bibliographystyle{ieeetr}
	\bibliography{myref}
}
\vfill

\end{document}